%% file: arxiv.tex
\documentclass[10pt,twocolumn,letterpaper]{article}

\usepackage[pagenumbers]{cvpr} %
\usepackage{xcolor}

\input{preamble}

\definecolor{cvprblue}{rgb}{0.21,0.49,0.74}
\usepackage[pagebackref,breaklinks,colorlinks,citecolor=cvprblue]{hyperref}

\def\paperID{2259} %
\def\confName{CVPR}
\def\confYear{2024}

\title{GAvatar: Animatable 3D Gaussian Avatars with Implicit Mesh Learning}

\begin{document}

\author{
Ye Yuan$^*$ \hspace{.5em} Xueting Li$^*$ \hspace{.5em} Yangyi Huang \hspace{.5em} Shalini De Mello \hspace{.5em} Koki Nagano \hspace{.5em} Jan Kautz \hspace{.5em} Umar Iqbal \\[1mm]
NVIDIA \\
{ \url{https://nvlabs.github.io/GAvatar}} \\
}

\twocolumn[{%
\renewcommand\twocolumn[1][]{#1}%
\maketitle
 \begin{center}
     \vspace{-7mm}
     \centering
     \includegraphics[trim={1cm 0 0 0}, width=1.0\textwidth]{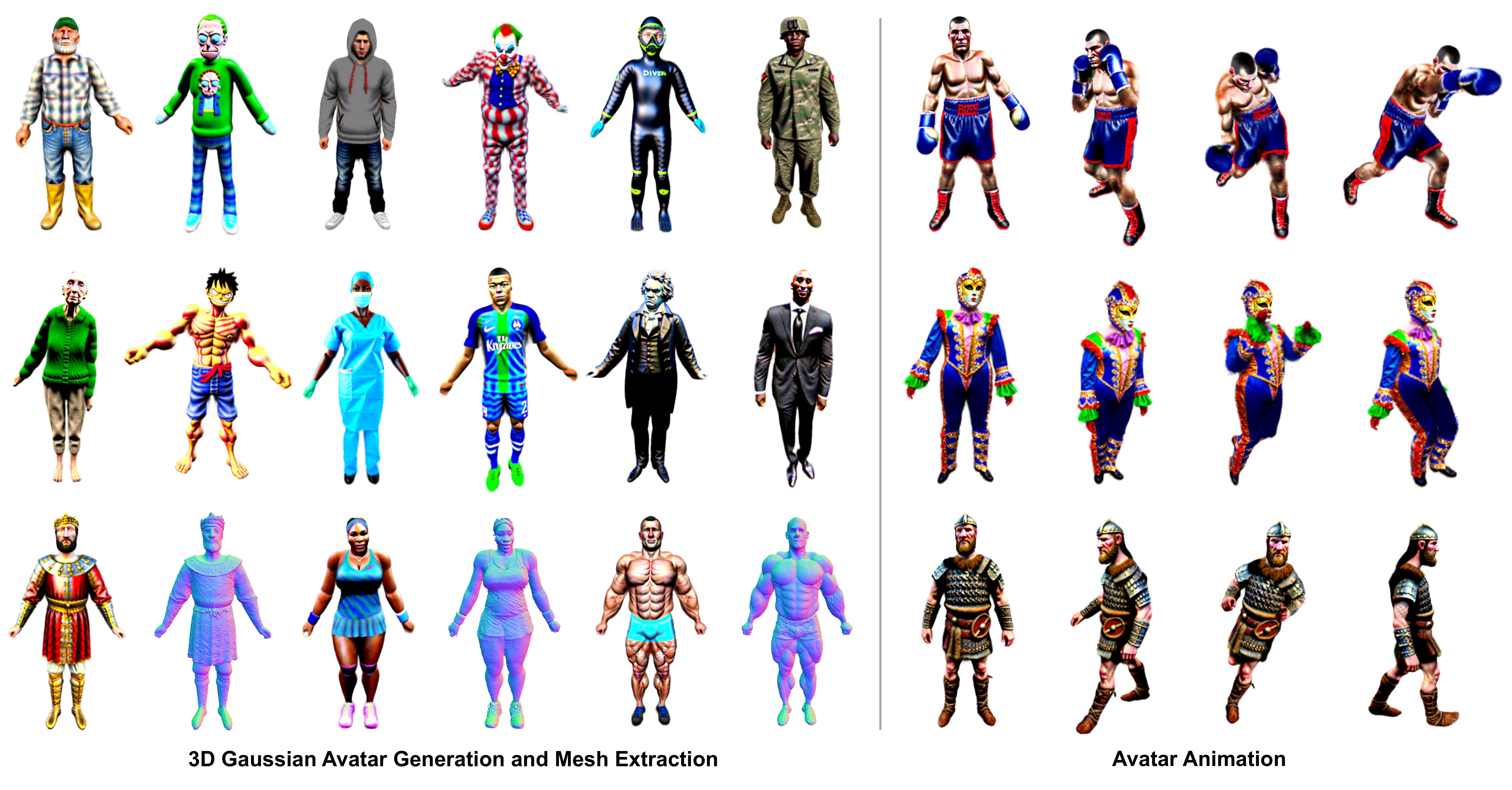}
     \vspace{-8mm}
     \captionof{figure}{GAvatar synthesizes high-fidelity 3D animatable avatars from text prompts. Our novel primitive-based implicit Gaussian representation enables efficient avatar animation (100 fps, 1K resolution) and also extracts a highly detailed mesh from learned 3D Gaussians.} %
     \vspace{0mm}
     \label{fig:teaser}
 \end{center}%
}]
\def\thefootnote{*}\footnotetext{Equal contribution.}\def\thefootnote{\arabic{footnote}}

\input{sec/0_abstract}    
\input{sec/1_intro}
\input{sec/2_relatedwork}

\input{sec/3_preliminaries}
\input{sec/4_approach}

\input{sec/5_experiments}
\input{sec/6_conclusion}
{
    \small
    \bibliographystyle{ieeenat_fullname}
    \bibliography{main}
}

\clearpage
\input{sec/supp_arxiv}

\end{document}

%% file: preamble.tex
\newcommand{\sdf}{\mathcal{S}_\psi}
\newcommand{\meshext}{\widetilde{\mathcal{M}}}

%% file: sec/0_abstract.tex
\begin{abstract}

\vspace{-2mm}
Gaussian splatting has emerged as a powerful 3D representation that harnesses the advantages of both explicit (mesh) and implicit (NeRF) 3D representations. In this paper, we seek to leverage Gaussian splatting to generate realistic animatable avatars from textual descriptions, addressing the limitations (e.g., flexibility and efficiency) imposed by mesh or NeRF-based representations. However, a naive application of Gaussian splatting cannot generate high-quality animatable avatars and suffers from learning instability; it also cannot capture fine avatar geometries and often leads to degenerate body parts. To tackle these problems, we first propose a primitive-based 3D Gaussian representation where Gaussians are defined inside pose-driven primitives to facilitate animation. Second, to stabilize and amortize the learning of millions of Gaussians, we propose to use neural implicit fields to predict the Gaussian attributes (e.g., colors). Finally, to capture fine avatar geometries and extract detailed meshes, we propose a novel SDF-based implicit mesh learning approach for 3D Gaussians that regularizes the underlying geometries and extracts highly detailed textured meshes. Our proposed method, GAvatar, enables the large-scale generation of diverse animatable avatars using only text prompts. GAvatar significantly surpasses existing methods in terms of both appearance and geometry quality, and achieves extremely fast rendering (100 fps) at 1K resolution.

\end{abstract}

%% file: sec/1_intro.tex
\vspace{-4mm}
\section{Introduction}
\label{sec:intro}
Digital avatars play an essential role in numerous applications, from augmented and virtual reality to gaming, movie production, and synthetic data generation
~\cite{zhu_2020_eccv_nba, deepcap, zheng2021deepmulticap, li2020monoport,li2020monoportRTL, zheng2023avatarrex, saito2020pifuhd, xiu2023econ}.
However, highly realistic and animatable avatars are extremely difficult to create at scale due to the complexity and diversity of character geometries and appearances. Traditional approaches rely on manual modeling and rigging of digital avatars, which are labor-intensive and time-consuming.
Recent advances in text-to-image generative models trained on large-scale data show impressive results in generating highly diverse and realistic human images from text~\cite{rombach2022high, saharia2022photorealistic,dalle2,zhang2023adding}.
In light of this, several methods are proposed to generate 3D avatars from textual descriptions by distilling the 2D prior of these generative models into 3D avatar representations~\cite{hong2022avatarclip, kolotouros2023dreamhuman, huang2023dreamwaltz}.
While their results are promising, the quality of the generated avatars is limited by the 3D representations they use, which are typically based on mesh or neural radiance field (NeRF)~\cite{mildenhall2020nerf}.
Mesh-based representations allow efficient rendering through rasterization, but the expressiveness to capture diverse geometry and fine details is limited due to the underlying topology. NeRF-based representations are expressive in modeling complex 3D scenes, but they are computationally expensive due to the large number of samples required by volume rendering to produce high-resolution images. 
As a result, existing avatar generation methods often fail to both generate fine-grained, out-of-shape geometric details, such as loose clothing, and efficiently render high-resolution avatars, which are critical for interactive and dynamic applications.

We aim to address these issues by adopting a new 3D representation, 3D Gaussian Splatting~\cite{kerbl3Dgaussians}, which represents a scene using a set of 3D Gaussians with color, opacity, scales, and rotations and produces rendering by differentiably splatting the Gaussians onto an image. Gaussian splatting combines the advantages of both mesh and NeRF-based representations and it is both efficient and flexible to capture fine details. 
However, naive applications of Gaussian splatting to avatar generation fail for several reasons due to the unconstrained nature of individual Gaussians. First, the Gaussian splatting representation is not animatable, as the Gaussians are defined in the world coordinate and cannot be easily transformed with the avatar's pose in a coherent manner. Second, a large number (millions) of Gaussians are required to model a highly detailed avatar, and the immense optimization space of individual Gaussian attributes (\eg, color, opacity, scale, rotation) leads to unstable optimization, especially when using high-variance objectives such as SDS~\cite{poole2022dreamfusion}. Third, the 3D Gaussians lack explicit knowledge of surfaces, and cannot easily incorporate surface normal supervision, which is crucial for extracting highly detailed 3D meshes~\cite{chen2023fantasia3d, huang2023humannorm}. Without geometry supervision, missing or degenerate body parts can appear when using weak 3D supervision (\ie, SDS), which we will show in the experiments.

To tackle these problems, we propose GAvatar, a novel approach that leverages Gaussian Splatting to generate realistic animatable avatars from textual descriptions.
First, we introduce a new primitive-based 3D Gaussian representation that defines 3D Gaussians inside pose-driven primitives. This representation naturally supports animation and enables flexible modeling of fine avatar geometry and appearance by deforming both the Gaussians and the primitives.
Second, we propose to use implicit Gaussian attribute fields to predict the Gaussian attributes, which stabilizes and amortizes the learning of a large number of Gaussians, and allows us to generate high-quality avatars using high-variance optimization objectives such as SDS. Additionally, after avatar optimization, since we can obtain the Gaussian attributes directly and skip querying the attribute fields, our approach achieves extremely fast (100 fps) rendering of neural avatars at a resolution of $1024{\times}1024$. This is significantly faster than existing NeRF-based avatar models \cite{kolotouros2023dreamhuman,cao2023dreamavatar} that query neural field for each novel camera view and avatar pose.
Finally, we also propose a novel signed distance function (SDF)-based implicit mesh learning approach that connects SDF with Gaussian opacities. Importantly, it enables GAvatar to regularize the underlying geometry of the Gaussian avatar and extract high-quality textured meshes.
\noindent Our contributions are summarized as follows:

\vspace{1mm}
\begin{itemize}[leftmargin=4.3mm]
\itemsep1mm 
    \item We introduce a new primitive-based implicit Gaussian representation for animatable avatars, enabling more stable and high-quality 3D avatar generation. It also allows extremely fast rendering (100 fps) at 1K resolution.
    \item We propose a novel SDF-based method that effectively regularizes the underlying geometry of 3D Gaussians and also enables the extraction of high-quality textured meshes from the learned Gaussians avatar.
    \item Our approach generates 3D avatars with fine geometry and appearance details. We experimentally demonstrate that GAvatar consistently outperforms existing methods in terms of avatar quality.
\end{itemize}

%% file: sec/2_relatedwork.tex
\section{Related Work}
\label{sec:related_work}
\noindent\textbf{3D Representations for 3D Content Generation.} 
Various 3D representations have been employed for 3D content generation, each with its own set of strengths and limitations. Triangulated meshes are a common choice due to their simplicity and compatibility with existing graphics pipelines~\cite{clipmatrix}. However, their inflexible topology can pose challenges in accurately representing intricate geometries. Alternatively, volumetric representations, such as voxel grids~\cite{sitzmann2019deepvoxels}, offer flexibility in modeling complex shapes. Nevertheless, their computational and memory costs grow cubically with resolution, impeding the faithful reconstruction of fine geometry details and smooth surfaces. Recently, NeRFs~\cite{mildenhall2020nerf} have gained prominence for modeling 3D shapes, especially in text-to-3D applications, thanks to their ability to capture arbitrary topologies with minimal memory usage. Yet, their rendering cost increases significantly at higher resolutions. Some approaches adopt hybrid representations to harness the benefits of different techniques. The Mixture of Volumetric Primitives (MVP) representation~\cite{Lombardi21}, for instance, introduces volumetric primitives onto a template mesh, achieving rapid rendering by leveraging a convolutional network to compute volumetric primitives. It generates images through ray-marching, accumulating colors and opacities from the primitives. Gaussian Splatting~\cite{kerbl3Dgaussians} has emerged as a promising 3D representation for efficiently rendering high-resolution images. It models objects using colored 3D Gaussians, which are rendered onto an image using splatting-based rasterization. However, a notable limitation is its difficulty in extracting meshes from learned Gaussians, as it predominantly captures appearance details through 3D Gaussians without modeling the underlying object surfaces. 

In this work. we introduce a novel primitive-based 3D Gaussian representation with implicit mesh learning.  It enables modeling dynamic and articulated objects like humans using Gaussian Splatting while also facilitating textured mesh extraction. In comparison to MVP, our Gaussian-based representation is more flexible and expressive, since each primitive comprises a variable number of 3D Gaussians with varying non-uniform locations that can go beyond the primitive boundaries. This allows it to capture finer details compared to the cubic primitives used in MVP. Moreover, our representation employs splatting-based rasterization, enabling efficient rendering of high-resolution images compared to traditional ray-marching techniques.

\vspace{2mm}
\noindent\textbf{Text-to-3D Generation.}
The field of text-to-3D generation has recently been revolutionized~\cite{poole2022dreamfusion, lin2022magic3d, richardson2023texture, chen2023fantasia3d, richardson2023texture, wang2023prolificdreamer, tang2023dreamgaussian} with the availability of large text-to-image models~\cite{rombach2022high, saharia2022photorealistic,dalle2,zhang2023adding}. The earlier methods optimize the 3D objects by encouraging the 2D rendering to be consistent with the input text in the CLIP~\cite{clip} embeddings space~\cite{jain2021dreamfields, sanghi2022clipforge, wang2022clipnerf, chen2022tango, xu2022dream3d, clipmatrix}. While they demonstrated the usefulness of text-to-image models for 3D content generation, the resulting 3D models often lacked realism and fine geometry details. The seminal work DreamFusion~\cite{poole2022dreamfusion} replaces the CLIP model with a text-to-image diffusion model and proposed Score Distillation Sampling (SDS) to optimize a NeRF-based representation of the 3D object. Since then multiple variants of this method have been proposed. 
Magic3D~\cite{lin2022magic3d} enhances runtime efficiency with a two-staged framework and adopts a more efficient DMTet~\cite{gao2020learning} representation. ProlificDreamer~\cite{wang2023prolificdreamer} addresses over-saturation/smoothing issues through a variational SDS objective. MVDream~\cite{shi2023MVDream} fine-tunes text-to-image models to generate 3D-consistent multi-view images, enabling efficient 3D generations. Fantasia3D~\cite{chen2023fantasia3d} disentangles geometry and appearance modeling, optimizing surface normals separately using the SDS loss. More recently, DreamGaussian~\cite{tang2023dreamgaussian} replaced the NeRF-based representation with Gaussian Splatting to significantly reduce runtime. However, this leads to 3D models with limited geometry and appearance quality, despite attempts to refine texture details through mesh-based fine-tuning. It is important to note that all these methods are limited to rigid objects only and cannot be animated easily. 

\vspace{2mm}
\noindent\textbf{Text-to-3D Avatar Generation}
Building upon the success achieved in generating static 3D objects, numerous methods have been proposed to model dynamic objects, particularly human or human-like avatars~\cite{clipmatrix, cao2023dreamavatar, kolotouros2023dreamhuman, jiang2023avatarcraft, zhang2023getavatar, huang2023dreamwaltz, zhang2023avatarverse, huang2023humannorm, huang2024tech, liao2024tada}. ClipMatrix~\cite{clipmatrix} is one of the first methods that showcased the creation of animatable avatars based on textual descriptions. It achieves this by optimizing a mesh-based representation using a CLIP-embedding loss. AvatarClip~\cite{hong2022avatarclip} follows a similar pipeline but employs a NeRF-based representation~\cite{wang2021neus}.
DreamAvatar~\cite{cao2023dreamavatar} and AvatarCraft~\cite{jiang2023avatarcraft} utilize SDS loss instead of CLIP, and learn the NeRF representation in canonical space through the integration of human body priors from SMPL~\cite{SMPL:2015}. DreamHumans~\cite{kolotouros2023dreamhuman} introduces a deformable and pose-conditioned NeRF model by incorporating the imGHUM~\cite{alldieck2021imghum} model. DreamWaltz~\cite{huang2023dreamwaltz} and AvatarVerse~\cite{zhang2023avatarverse} leverage pose-conditioned ControlNets~\cite{zhang2023adding}, showcasing improved avatar quality with conditional SDS.
However, a common limitation among these methods is their reliance on NeRF to generate images, resulting in the computation of SDS loss based on low-resolution images. For instance, DreamHumans~\cite{kolotouros2023dreamhuman} generates $64{\times}64$ images during optimization, leading to a compromise in avatar quality. In contrast, our approach can efficiently generate images with a resolution of $1024{\times}1024$, resulting in higher-quality avatars, as demonstrated in our experiments.
There are several contemporary works that demonstrate impressive avatar quality~\cite{huang2023humannorm, Zeng2023AvatarBoothHA, liao2024tada}. TADA~\cite{liao2024tada} shows that a mesh-based approach with adaptive mesh subdivision can be used to generate high-quality avatars. HumanNorm~\cite{huang2023humannorm} finetunes text-to-image models to directly generate normal and depth maps from the input text. The adapted models are then utilized to optimize the avatar's geometry through the SDS loss, with texture optimization achieved using a normal-conditioned ControlNet~\cite{zhang2023adding}. Similarly,  AvatarBooth~\cite{Zeng2023AvatarBoothHA} fine-tunes region-specific diffusion models, highlighting that employing dedicated models for distinct body regions enhances avatar quality. These improved optimization objectives are complementary to our method since they are compatible with our Gaussian-based 3D representation. Since our model can efficiently render high-resolution images and normals, we anticipate synergies between our approach and~\cite{huang2023humannorm, Zeng2023AvatarBoothHA, zhang2023avatarverse} to yield further enhancements.

%% file: sec/3_preliminaries.tex
\begin{figure*}[t]
    \centering
    \includegraphics[width=1.0\textwidth]{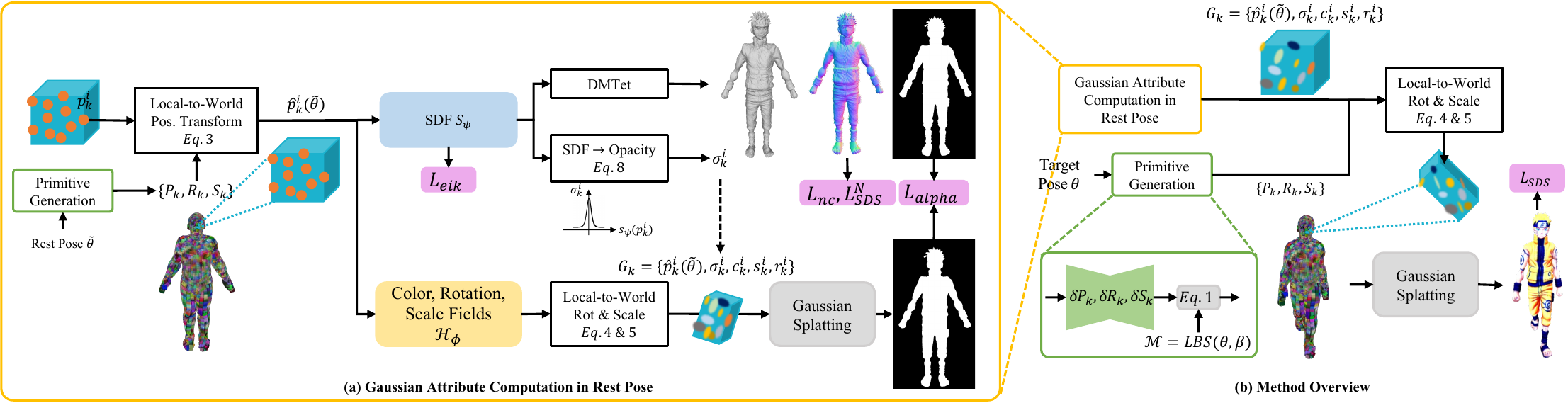}
    \vspace{-6mm}
    \caption{\textbf{Overview of GAvatar.} We first generate the primitives $V_k{=}(P_k, R_k, S_k)$ in the rest pose $\tilde{\theta}$. Each primitive consists of $N_k$ 3D Gaussians with their position $p_k^i$, rotation $r_k^i$ and scaling $s_k^i$ defined in the primitive's local coordinate system. Next, we obtain the canonical positions, $\hat{p}^i_k(\tilde{\theta})$, of the Gaussians in the world coordinates by applying the global transforms of the primitives using Eq.~\ref{eq:gaussian}.  These positions are then used to query the color $c_k^i$, rotation $r_k^i$ and scaling $s_k^i$ of each Gaussian from a neural attribute field $\mathcal{H}_\phi$. Each Gaussian's SDF value is queried from a neural SDF $\mathcal{S}_\psi$ and is converted into the opacity $\sigma_k^i$ through a kernel function $\mathcal{K}$. The 3D Gaussians with the predicted attributes are then rasterized onto the camera view using Gaussian splatting to produce the RGB image $I$ and alpha image $I_\alpha$. We use DMTet~\cite{shen2021deep} to differentiably extract the mesh from the Gaussian SDF values and generate its normal map and silhouette for geometry regularization. For animating the avatar using any target pose $\theta$, we generate the primitives using the target pose and use them to transform the 3D Gaussians, before rasterizing the image. A method walkthrough is also provided in the supplementary \href{https://youtu.be/PbCF1HzrKrs}{video}.}
    \label{fig::overview}
    \vspace{-3mm}
\end{figure*}

\section{Preliminaries}
\label{sec: preliminaries}

\noindent\textbf{Primitive-based 3D Representation.}
Primitive-based methods represent a 3D scene by a set of primitives such as cubes~\cite{Lombardi21,remelli2022drivable}, points~\cite{aliev2020neural} or nerflets~\cite{zhang2022nerfusion}. 
In this work, we adopt the primitive formulation used in~\cite{Lombardi21,remelli2022drivable}: a set of $K$ cubic primitives $\{V_1,\ldots,V_K\}$ are attached to the surface of a SMPL-X~\cite{SMPL-X:2019} mesh $\mathcal{M} = \text{LBS}(\theta,\beta)$, where $\theta$ and $\beta$ are the SMPL-X pose and shape parameters, and LBS is the linear blend skinning function. Each primitive $V_k=\{P_k, R_k, S_k\}$ is defined by its location $P_k\in \mathbb{R}^3$, per-axis scale $S_k\in \mathbb{R}^3$ and orientation $R_k\in \text{SO}(3)$. 
The primitive parameters are generated by:
\begin{equation}
    \label{eq:primitive}
    \begin{split}
        P_k(\theta) &= \hat{P}_k(\mathcal{M}) + \delta P_\omega(\theta)_k, \\ 
        R_k(\theta) &= \delta R_\omega(\theta)_k \cdot \hat{R}_k(\mathcal{M}), \\ 
        S_k(\theta) &= \hat{S}_k(\mathcal{M}) + \delta S_\omega(\theta)_k,
    \end{split}
\end{equation}
where we first compute a mesh-based primitive initialization $\hat{P}_k(\mathcal{M}), \hat{R}_k(\mathcal{M}), \hat{S}_k(\mathcal{M})$, and then apply pose-dependent correctives $\delta P_\omega(\theta), \delta R_\omega(\theta), \delta S_\omega(\theta)$, which are represented by neural networks with parameters $\omega$. The mesh-based initialization is computed by placing the primitives on a 2D grid in the mesh's $uv$-texture space and generating the primitives at the 3D locations on the mesh surface points corresponding to the $uv$-coordinates. The overall deformation process is illustrated in Fig.~\ref{fig::overview} (green box) and more details can be found in~\cite{remelli2022drivable}.

\vspace{2mm}\noindent\textbf{Score Distillation Sampling.}
First proposed in DreamFusion~\cite{poole2022dreamfusion}, score distillation sampling (SDS) can be used to optimize the parameters $\eta$ of a 3D model $g$ using a pretrained text-to-image diffusion model. Given a text prompt $y$ and the noise prediction $\hat{\epsilon}(I_t; y, t)$ of the diffusion model, SDS optimizes model parameters $\eta$ by minimizing the difference between the noise $\epsilon$ added to the rendered image $I = g(\eta)$ and the predicted noise $\hat{\epsilon}$ by the diffusion model:
\begin{equation}
\label{eq:sds}
    \nabla_{\eta} \mathcal{L}_\text{SDS} = E_{t,\epsilon} \left[w(t)(\hat{\epsilon}(I_t; y, t) - \epsilon) \frac{\partial I}{\partial \eta} \right]\,,
\end{equation}
where $g(\eta)$ denotes the differentiable rendering process of the 3D model, $t$ is the noise level, $I_t$ is the noised image, and $w(t)$ is a weighting function.

%% file: sec/4_approach.tex
\section{Approach}
\label{sec:approach}

Our approach, GAvatar, generates a 3D Gaussian-based animatable avatar given a text prompt. Our key ideas are two-fold: (1) we introduce a new primitive-based implicit 3D Gaussian representation (Sec.~\ref{sec:representation}) that not only enables avatar animation but also stabilizes and amortizes the learning of a large number of Gaussians using the high-variance SDS loss; (2) we represent the underlying geometry of 3D Gaussians with an SDF that enables extracting high-quality textured meshes and regularizing the avatar's geometry (Sec.~\ref{sec:sdf}). The training process of our approach is described in Sec.~\ref{sec:optimization} and an overview of our method is provided in Fig.~\ref{fig::overview}.

\subsection{Primitive-based Implicit Gaussian Avatar}
\label{sec:representation}
Recently, Gaussian Splatting~\cite{kerbl3Dgaussians} has emerged as a powerful representation for 3D scene reconstruction and generation thanks to its efficiency and flexibility.
However, naive application of Gaussian Splatting to human avatar generation poses animation and training stability challenges.
Specifically, two essential questions arise: (1) how to transform the Gaussians defined in the world coordinate system along with the deformable avatar and (2) how to learn Gaussians with consistent attributes (i.e., color, rotation, scaling, etc.) within a local neighborhood.
In the following, we answer both questions by proposing a primitive-based implicit Gaussian representation.

\vspace{2mm}\noindent\textbf{Primitive-based 3D Gaussian Avatar.}
To generate an animatable human avatar, we start with the primitive formulation discussed in Sec.~\ref{sec: preliminaries}, where the human body is represented by a set of primitives attached to its surface.
Since the primitives are naturally deformed according to the human pose and shape, we propose to attach a set of 3D Gaussians $\{G_k^1, \ldots, G_k^{N_k}\}$ to the local coordinate system of each primitive $V_k{=}\{P_k, R_k, S_k\}$ and deform them along with the primitive. Specifically, each Gaussian $G_k^i{=}\{p_k^i, r_k^i, s_k^i, c_k^i, \sigma_k^i\}$ is defined by its position $p_k^i$, rotation $r_k^i$, and scaling $s_k^i$ in the primitive's local coordinates, as well as its color features $c_k^i$ and opacity $\sigma_k^i$.
Given a target pose $\theta$, we first obtain the location $P_k$, scale $S_k$, and orientation $R_k$ of each deformed primitive using Eq.~\ref{eq:primitive}. Then the global location $\hat{p}_k^i$, scale $\hat{s}_k^i$, and orientation $\hat{r}_k^i$ of each Gaussian $G_k^i$ associated with the primitive are computed as:
\begin{align}
\label{eq:gaussian}
    \hat{p}_k^i(\theta) &= R_k(\theta)\cdot (S_k(\theta)\odot p_k^i) + P_k(\theta)\\
    \hat{s}_k^i(\theta) &= S_k(\theta)\cdot s_k^i \\
    \hat{r}_k^i(\theta) &= R_k(\theta)\cdot r_k^i
\end{align}
This primitive-based Gaussian representation naturally balances constraint and flexibility. It is more flexible compared to the native primitive representation in~\cite{Lombardi21,remelli2022drivable} since it allows a primitive to deform beyond a cube by equipping it with Gaussians. Meanwhile, the Gaussians within each primitive share the motion of the primitive and are more constrained during animation. 

\vspace{2mm}\noindent\textbf{Implicit Gaussian Attribute Field.}
To fully exploit the expressiveness of 3D Gaussians, we allow each Gaussian to have individual attributes, \ie, color features, scaling, rotation, and opacity. However, this potentially results in unstable training where Gaussians within a local neighborhood possess different attributes, leading to noisy geometry and rendering. This is especially true when the gradient of the optimization objective has high variance, such as the SDS objective in Eq.~\ref{eq:sds}.
To stabilize and amortize the training process, instead of directly optimizing the attributes of the Gaussians, we propose to predict these attributes using neural implicit fields.
As shown in the yellow block in Fig.~\ref{fig::overview}, for each Gaussian $G_k^i$, we first compute its canonical position $\hat{p}_k^i(\tilde{\theta})$ in the world coordinate system (Eq.~\ref{eq:gaussian}), where $\tilde{\theta}$ represents the rest pose. We can then query the color $c_k^i$, rotation $r_k^i$, scaling $s_k^i$ and opacity $\sigma_k^i$ of each Gaussian using the canonical position $\hat{p}_k^i(\tilde{\theta})$ from two neural implicit fields $\mathcal{H}_\phi$ and $\mathcal{O}_\psi$, which are represented by neural networks with parameters $\phi$ and~$\psi$:
\begin{align}
\label{eq:field}
    (c_k^i, r_k^i, s_k^i) & = \mathcal{H}_\phi(\hat{p}_k^i(\tilde{\theta})) \\
    \sigma_k^i & = \mathcal{O}_\psi(\hat{p}_k^i(\tilde{\theta}))
\end{align}
where we use a separate neural field $\mathcal{O}_\psi$ to output the opacities of the Gaussians, while other attributes are predicted by $\mathcal{H}_\phi$. This design is because the opacities of the Gaussians are closely related to the underlying geometry of the avatar and require special treatment, which will be discussed in Sec.~\ref{sec:sdf}.
Note that by querying the neural field with a canonical rest pose $\tilde{\theta}$, we canonicalize the Gaussian attributes, which can then be shared across different poses and animations.
Our use of neural implicit fields constrains nearby Gaussians to have consistent attributes, which greatly stabilizes and amortizes the training process and enables high-quality avatar synthesis using high-variance losses.

\vspace{2mm}\noindent\textbf{Rendering and Objectives.}
After obtaining the positions and attributes of 3D Gaussians, we adopt the efficient Gaussian splatting technique described in~\cite{kerbl3Dgaussians} to render an RGB image $I$ and also an alpha image $I_\alpha$. The RGB image $I$ is then used for the SDS loss defined in Eq.~\ref{eq:sds} as one of the main training objectives.
To prevent the Gaussians from straying far away from the primitives, we also utilize a local position regularization loss $\mathcal{L}_\text{pos} {=}\sum_{k,i}\|p_k^i\|^2$, which constrains the Gaussians to be close to the origin of the associated primitives.

\subsection{SDF-based Mesh Learning for 3D Gaussians}
\label{sec:sdf}
    
A crucial aspect yet to be addressed in our primitive-based 3D Gaussian representation is how to properly represent the underlying geometry of the 3D Gaussians.
This is important for two reasons: (1) 3D Gaussians are transparent ``point clouds'' that do not have well-defined surfaces, which can lead to degenerate body parts or holes in the generated avatars (see Fig.~\ref{fig::ablation}); (2) Currently, there is no efficient and effective way to extract textured meshes from a large number of 3D Gaussians, which are often important for applications in traditional graphics pipelines.

\vspace{2mm}\noindent\textbf{SDF-based Gaussian Opacity Field.}
To address this problem, we propose to represent the underlying geometry of 3D Gaussians through a signed distance field (SDF) function $\sdf$ with parameters $\psi$. Specifically, we parametrize the opacity $\sigma_k^i$ of each 3D Gaussian based on their signed distance to the surface using a kernel function $\mathcal{K}$ inspired by NeuS~\cite{wang2021neus}:
\begin{equation}
\label{eq:sdf}
\begin{split}
    \sigma_k^i = \mathcal{K}(\sdf(p_k^i))\,,
\end{split}
\end{equation}
where $\mathcal{K}(x) = \gamma e^{-\lambda x}/(1 + e^{-\lambda x})^2 $ is a bell-shaped kernel function with learnable parameters $\{\gamma, \lambda\}$ that maps the signed distance to an opacity value. Intuitively, this opacity parametrization builds in the prior that Gaussians should stay close to the surface in order to obtain high opacity. The parameter $\lambda$ controls the tightness of the high-opacity neighborhood of the surface and $\alpha$ controls the overall scale of the opacity. The SDF-based Gaussian opacity parametrization naturally fits our primitive-based implicit Gaussian representation, since now we can define the aforementioned opacity field $\mathcal{O}_\psi$ as the product of the SDF and the kernel function: $\mathcal{O}_\psi = \mathcal{K} \circ \sdf$, and we can directly use a neural network to represent the SDF $\sdf$.

\vspace{2mm}\noindent\textbf{Mesh Extraction and Geometry Regularization.}
An important advantage of using an SDF $\sdf$ to represent the underlying geometry of 3D Gaussians is that it allows us to extract a mesh $\meshext$ from the SDF through differentiable marching tetrahedra (DMTet \cite{shen2021deep}):
\begin{equation}
    \meshext = \textsc{DMTet}(\sdf)\,.
\end{equation}
Both the SDF and extracted mesh allow us to utilize various losses to regularize the geometry of the 3D Gaussian avatar. Specifically, we first employ an Eikonal regularizer to maintain a proper SDF, which is defined as:
\begin{equation}
    \mathcal{L}_\text{eik} = (\|\nabla_{p} \sdf(p)\| - 1)^2\,,
\end{equation}
where $p \in \mathcal{P}$ contains both the center points of all Gaussians in the world coordinates as well as points sampled around the Gaussians using a normal distribution. Next, we also employ an alpha loss to match the mask $I_M$ rendered using the extracted mesh to the alpha image $I_\alpha$ from the Gaussian splatting:
\begin{equation}
    \mathcal{L}_\text{alpha} = \|I_M - I_\alpha\|^2\,.
\end{equation}
Inspired by Fantasia3D~\cite{chen2023fantasia3d}, we also use a normal SDS loss to supervise the normal rendering $I_N$ of the extracted mesh using differentiable rasterization~\cite{Laine2020diffrast}. The SDS gradient can be computed as:
\begin{equation}
    \nabla_{\theta} \mathcal{L}^N_\text{SDS} = E_{t,\epsilon} \left[w(t)(\hat{\epsilon}(I_{N,t}; y, t) - \epsilon) \frac{\partial I_N}{\partial \theta} \right]\,,
\end{equation}
where $I_{N,t}$ is the noised normal image. We further use a normal consistency loss $\mathcal{L}_\text{nc}$ which regularizes the difference between the adjacent vertex normals of mesh $\meshext$.

\vspace{2mm}\noindent\textbf{Texture Extraction.}
Our proposed implicit Gaussian attribute field $\mathcal{H}_\phi$ naturally facilitates texturing the extracted mesh $\meshext$, since we can use the Gaussian color field as the 3D texture field used by the differentiable rasterization. Once the Gaussian-based avatar is fully optimized, directly using the Gaussian color field already provides a good initial texture for the mesh, but we can further improve the texture quality by finetuning the color field using an SDS loss $\mathcal{L}_\text{sds}^{\meshext}$ on the RGB rendering $I_{\meshext}$ of the textured mesh. We observe that only a small number of finetuning iterations is required for convergence.

\subsection{Optimization}
\label{sec:optimization}
The overall objective of our method can be summarized as:
\begin{equation}
    \mathcal{L} = \mathcal{L}_\text{SDS} + \mathcal{L}_\text{pos} + \mathcal{L}_\text{eik} + \mathcal{L}_\text{alpha} + \mathcal{L}^N_\text{SDS} + \mathcal{L}_\text{nc}\,,
\end{equation}
where we omit the weighting terms for brevity. Using this objective, we optimize the Gaussian local positions $\{p_k^i\}$, Gaussian attribute field $\mathcal{H}_\phi$ and SDF $\mathcal{S}_\psi$, opacity kernel parameters $\{\gamma, \lambda\}$, 
primitive motion corrective networks $\delta P_\omega, \delta R_\omega, \delta S_\omega$, as well as the SMPL-X shape parameters~$\beta$.

\vspace{2mm}\noindent\textbf{Initialization.}
We divide the $uv$-map of SMPL-X into a $64\times 64$ grid, which gives us 4096 primitives. We assign 64 Gaussians to each primitive $V_k$ and initialize their local positions $\{p_k^i\}$ with a uniform grid of $4\times4\times4$.

\vspace{2mm}\noindent\textbf{Training.}
We perform Gaussian densification as described in~\cite{kerbl3Dgaussians} every 100 iterations, which leads to different numbers of Gaussians per primitive. We stop densification when the total number of Gaussians exceeds 2 million. To render the RGB image $I$ for the SDS loss $\mathcal{L}_\text{sds}$, we take the target pose $\theta$ from two sources: (1) a natural pose $\theta_N$ optimized together with the aforementioned variables; (2) a random pose $\theta_A$ sampled from an animation database to ensure realistic animation.

%% file: sec/5_experiments.tex
\vspace{-2mm}
\section{Experiments}
\label{sec:exp}
\vspace{-1mm}
\begin{figure*}[t]
    \centering
    \includegraphics[width=0.99\textwidth]{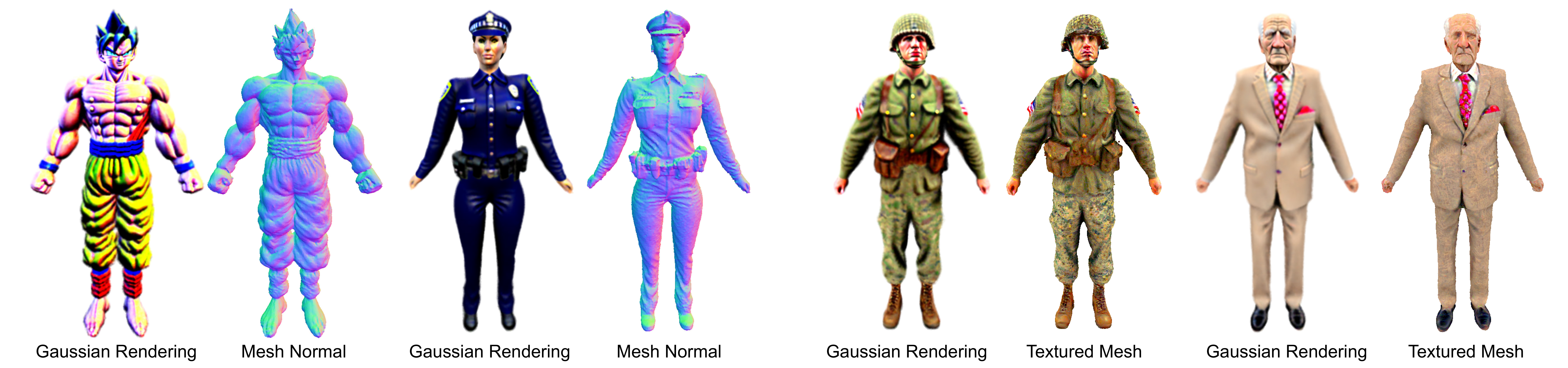}
    \vspace{-1mm}
    \caption{\textbf{Generated avatars by our method and their mesh normals and texture meshes.}}
    \label{fig:avatars}
    \vspace{-3mm}
\end{figure*}

\begin{figure*}[t]
    \centering
    \includegraphics[width=0.99\textwidth]{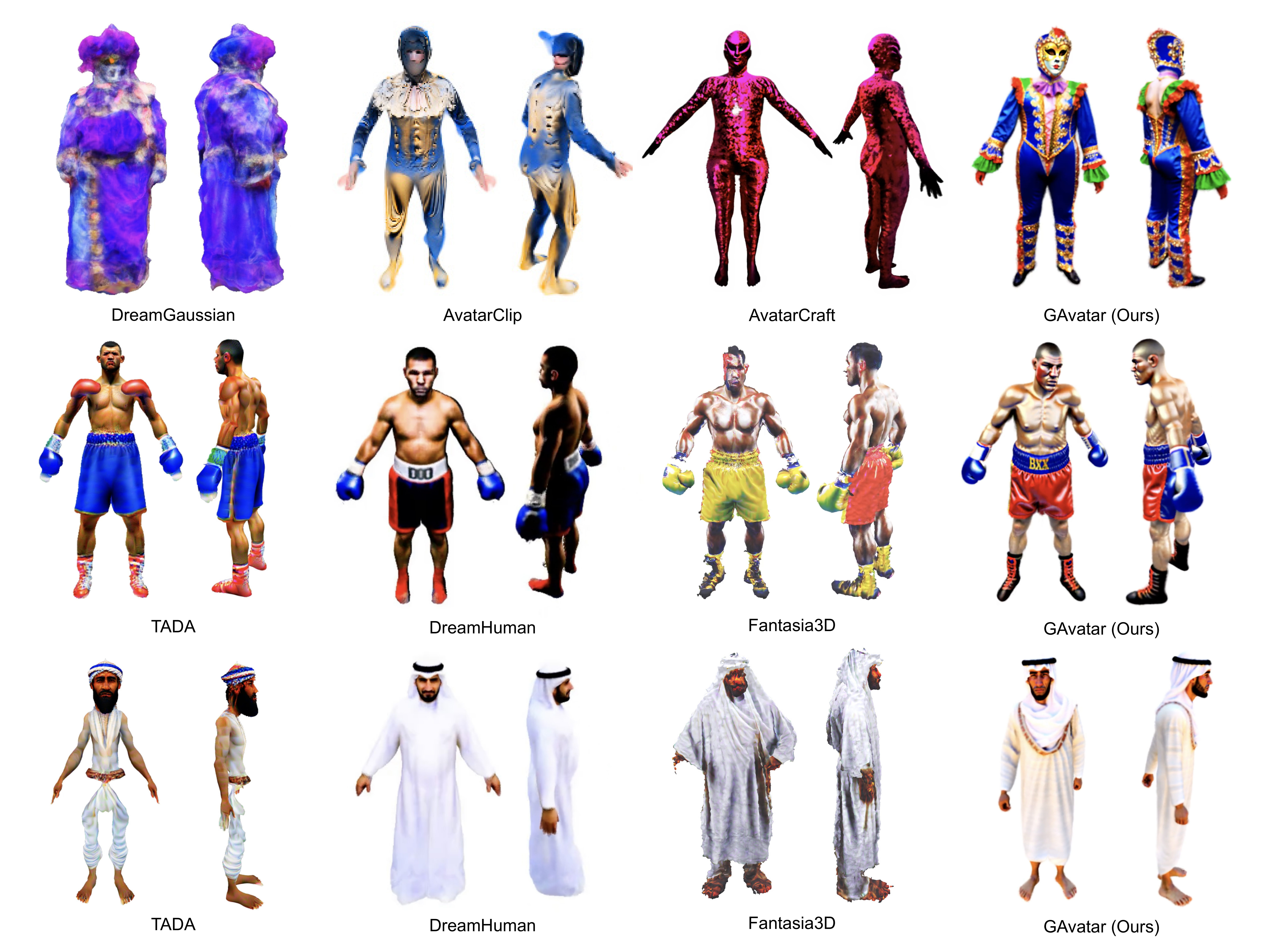}
    \vspace{-3mm}
    \caption{\textbf{Comparison with the state-of-the-art methods.}
    From top to bottom, the prompts used in each row are ``a person dressed at the venice carnival'', ``a professional boxer'' and ``a bedouin dressed in white''. Our method consistently produces the best quality avatars.}
    \label{fig:baseline_comparison}
    \vspace{-3mm}
\end{figure*}

In Fig.~\ref{fig:avatars}, we showcase example avatars generated by our method and their geometry and textured meshes. Notice the intricate geometry details captured by our method, thanks to our SDF-based implicit mesh learning for 3D Gaussians. Due to its primitive-based design, our approach readily supports avatar animation. We showcase various animations in Fig.~\ref{fig:teaser} and on the project \href{https://nvlabs.github.io/GAvatar}{website}.

\vspace{2mm}\noindent\textbf{Rendering Efficiency.}
Since GAvatar no longer needs to query the Gaussian attributes from the implicit fields after optimization, it achieves extremely fast rendering speed due to the use of 3D Gaussians. Specifically, a generated avatar with 2.5 million Gaussians can be rendered with $1024{\times}1024$ resolution at 100 fps, which is tremendously faster than most NeRF-based approaches. Moreover, the Gaussian rendering only takes about 3ms ($300+$ fps), so further speedup is possible by optimizing the speed of non-rendering operations such as LBS and primitive transforms.

\subsection{Qualitative Evaluation}
Fig.~\ref{fig:baseline_comparison} compares our method, GAvatar, with the state-of-the-art approaches: DreamGaussian~\cite{tang2023dreamgaussian}, AvatarCLIP~\cite{hong2022avatarclip}, AvatarCraft~\cite{jiang2023avatarcraft} and Fantasia3D~\cite{chen2023fantasia3d}. For completeness, we also compare with contemporary works, DreamHumans~\cite{kolotouros2023dreamhuman} and TADA~\cite{liao2024tada}. For DreamHumans~\cite{kolotouros2023dreamhuman} we use the avatar renderings provided on the project page, while for other methods we use the publicly available source codes. 
Our method clearly produces higher-quality avatars both in terms of geometry and appearance. DreamGaussian~\cite{tang2023dreamgaussian}, AvatarCLIP~\cite{hong2022avatarclip}, AvatarCraft~\cite{jiang2023avatarcraft} and Fantasia3D~\cite{chen2023fantasia3d} fail catastrophically to model complex avatars. DreamHumans~\cite{kolotouros2023dreamhuman} creates low-resolution avatars since it is trained with a resolution of $64{\times}64$ only. TADA~\cite{liao2024tada} can render high-resolution images due to a mesh-based rendering but can produce degenerate solutions with implausible shapes. It also provides smoother texture and less geometry details as compared to our method. GAvatar generates significantly better avatars as compared to all methods as we will also show in our user study next.
 
\subsection{Quantitative Evaluation}
To quantitatively evaluate the proposed method, we follow previous works~\cite{liao2024tada,hong2022avatarclip,tang2023dreamgaussian} and carry out an extensive A/B user study. We adopt 24 prompts commonly used in the baselines to generate the avatars. In total, we collected 1512 responses from 42 participants. For each vote, we show a pair of randomly chosen 3D avatars synthesized by our method and one of the baseline methods. We ask the participant to choose the method that has better 1) geometry quality, 2) appearance quality, and 3) consistency with the given prompt. Table~\ref{tab::user_study} summarizes the preference percentage of our method over the baseline methods. Notably, our method consistently outperforms existing and contemporary methods by a substantial margin.

\begin{table}[t]
\centering
\footnotesize
\resizebox{\linewidth}{!}{ 
\begin{tabular}{l|ccc}
\hline
Compared Method & Geometry Quality & Appearance Quality & Consistency with Prompt \\
\hline
AvatarCLIP~\cite{hong2022avatarclip} & 98.81 & 97.62 & 97.62 \\
AvatarCraft~\cite{jiang2023avatarcraft} & 96.43 & 98.81 & 98.81 \\
DreamGaussian~\cite{tang2023dreamgaussian} & 100.0 & 98.81 & 98.81\\
Fantasia3D~\cite{chen2023fantasia3d} & 92.86 & 92.86 & 91.67 \\
DreamHuman~\cite{kolotouros2023dreamhuman}* & 73.81 & 73.81 & 65.48\\
TADA~\cite{liao2024tada}* & 61.90 & 69.05 & 67.86\\
\hline
\end{tabular}
}
\vspace{-2mm}
\caption{\textbf{User Study.} We show a \textit{preference percentage} of our method over state-of-the-art methods (* denotes contemporary methods). GAvatar is preferred by the users over all baselines. }
\label{tab::user_study}
\end{table}

\begin{figure}[t]
    \centering
    \includegraphics[width=0.99\linewidth]{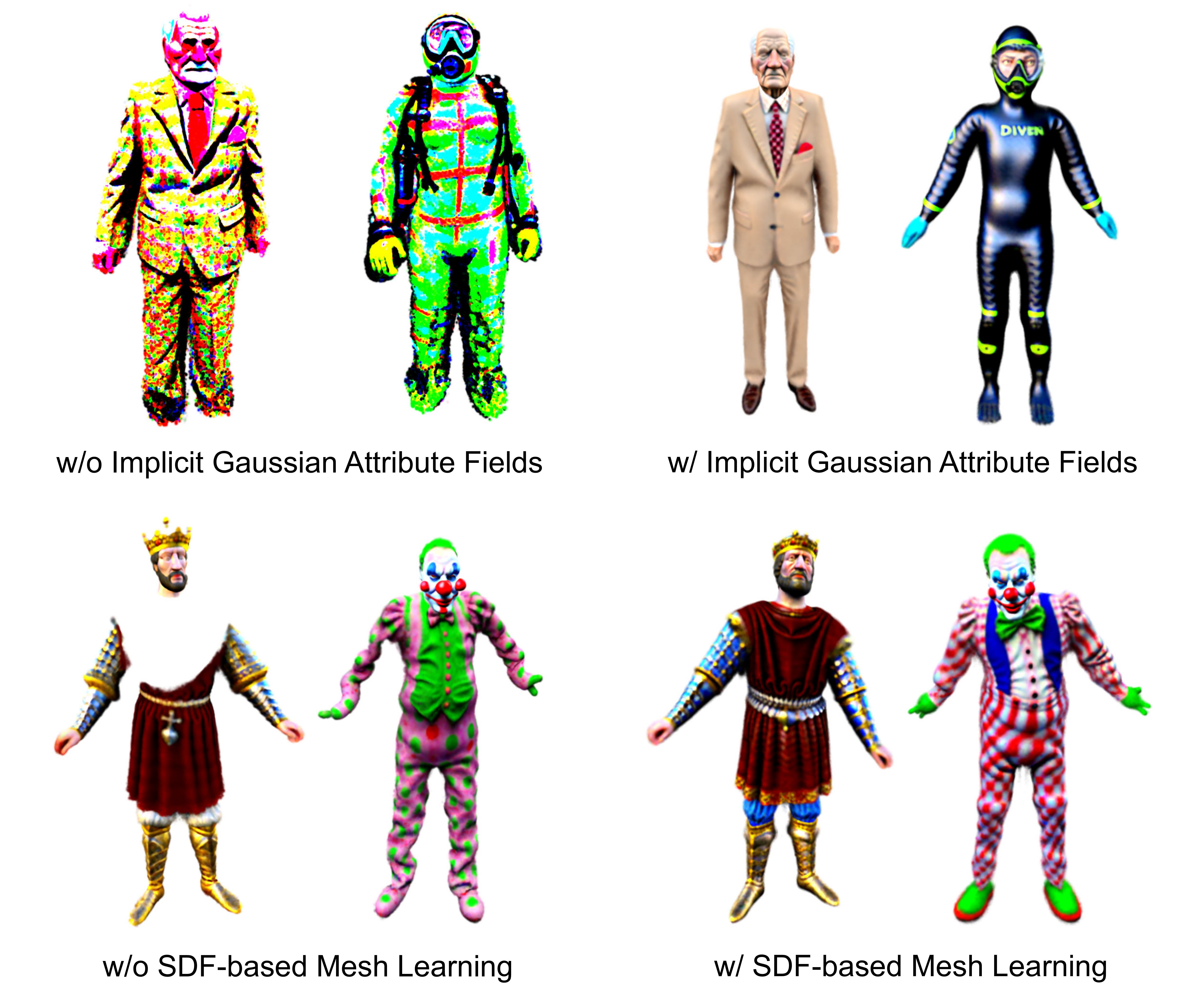}
    \vspace{-2mm}
    \caption{\textbf{Ablation Studies.}}
    \vspace{-3mm}
    \label{fig::ablation}
\end{figure}

\begin{figure}[t]
    \centering
    \includegraphics[width=0.99\linewidth]{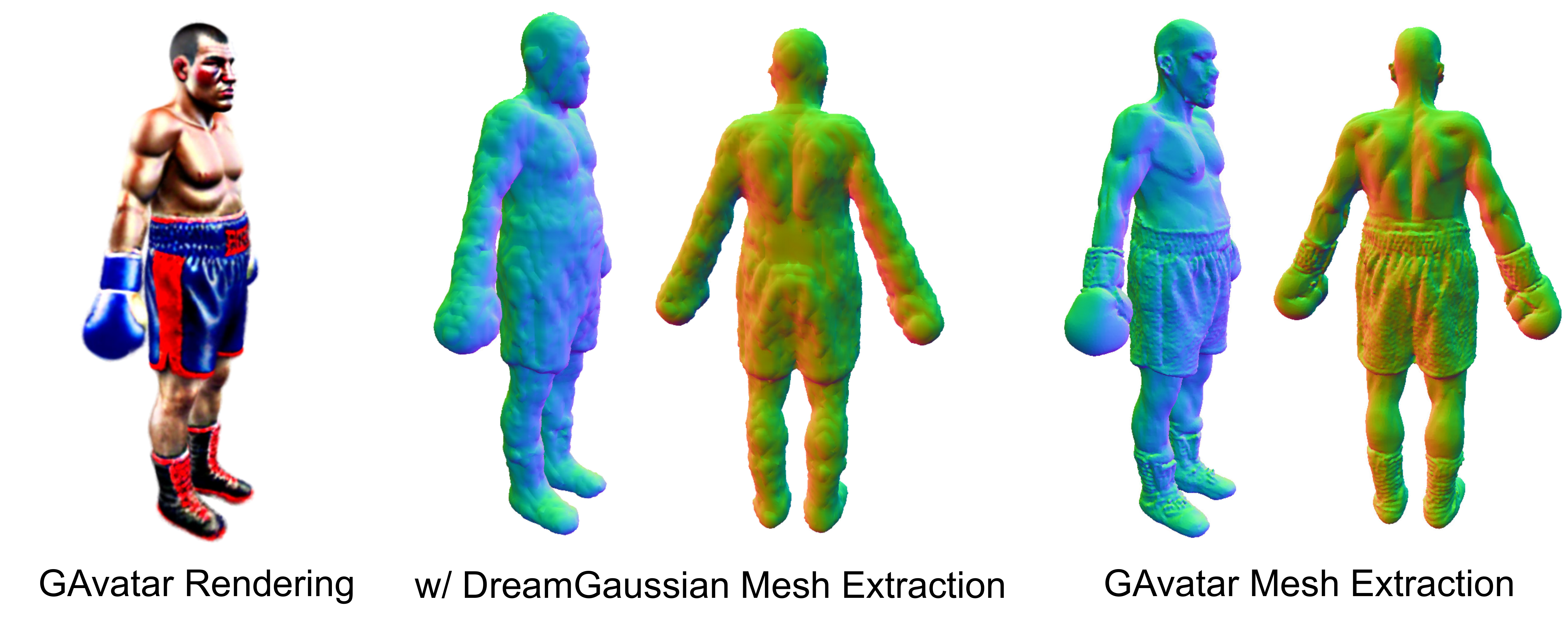}
    \vspace{-2mm}
    \caption{\textbf{Mesh Extraction Comparison.}}
    \vspace{-3mm}
    \label{fig::ablation_mesh}
\end{figure}

\subsection{Ablation Study}

\noindent\textbf{Effect of Implicit Gaussian Attribute Field.}
In Fig.~\ref{fig::ablation} (Top), we design a variant of our method by disabling the implicit Gaussian attribute field and directly optimizing the Gaussian attributes. We observe that the generated avatars are significantly worse than our method, with pronounced noise and color oversaturation. This aligns with our intuition that directly optimizing millions of Gaussians individually with high-variance loss like SDS is quite challenging. In contrast, our implicit Gaussian attribute field allows a much more stable and robust optimization process.

\vspace{2mm}\noindent\textbf{Effect of SDF-based Mesh Learning.}
In Fig.~\ref{fig::ablation} (Bottom), we design a variant of our approach by disabling the SDF-based mesh learning and instead letting the Gaussian attribute field additionally output the Gaussian opacities. As shown in Fig.~\ref{fig::ablation}, the generated avatars without mesh learning can have missing body parts and distorted body shapes. Our SDF-based mesh learning tackles these issues by regularizing the underlying geometry of the Gaussian avatar.

\vspace{2mm}\noindent\textbf{Mesh Extraction Comparison.}
An important benefit of our approach is that it allows us to extract a high-quality differentiable mesh representation of the Gaussian avatar. We compare our mesh extraction approach with the Gaussian density-based approach used in DreamGaussian~\cite{tang2023dreamgaussian}, one of the few works that extract meshes from 3D Gaussians. In particular, we provide its mesh extraction pipeline with our optimized Gaussian attributes to obtain the final mesh. The results are shown in Fig.~\ref{fig::ablation_mesh}. We observe the mesh extracted by DreamGaussian is more noisy and lacks geometry details, while our approach obtains much smoother meshes with fine-grained geometry details.

%% file: sec/6_conclusion.tex
\section{Discussion and Limitations}
\label{sec:conclusion}

We have presented a novel approach for generating diverse and animatable avatars with geometry learning and regularization. Our primitive-based 3D Gaussian representation allows us to flexibly model avatar geometry and appearance while enabling animation with extremely fast rendering. We demonstrated our neural implicit Gaussian attribute fields stabilize the learning of millions of 3D Gaussian under noisy objectives. We further propose a novel SDF-based mesh learning approach that regularizes the underlying geometry of the Gaussian avatar and extracts a high-quality textured mesh from 3D Gaussians. Our experiments and user study indicate that our approach surpasses state-of-the-art methods in terms of appearance and geometry quality.

While our approach has shown promising results, it still has several limitations to be addressed in future work. First, similar to other SDS-based approaches, our method sometimes also suffers from color oversaturation. We believe that exploring various techniques for improving SDS~\cite{wang2023prolificdreamer,lin2023common,katzir2023noise} can help mitigate this issue. Second, there can still be misalignment between the geometry and appearance of the generated avatars, where some geometry details in the rendering are embedded in the colors of the 3D Gaussians, similar to how texture can embed geometry details in mesh-based rendering. We believe that having consistent geometry and appearance supervisions such as those in HumanNorm~\cite{huang2023humannorm} can help alleviate this issue. Disentangling lighting and appearance details within the 3D Gaussian-based representation is also an interesting future direction. Lastly, animating loose clothing with correct temporal deformations is still challenging, especially when no direct image or temporal supervision is provided. Leveraging temporal priors such as physics simulation or video diffusion models can be a promising future avenue to explore.

%% file: sec/supp_arxiv.tex
\appendix

\section{Implementation Details}
\label{sec::supp_implementation_details}
\paragraph{Camera sampling.}
During optimization, we randomly sample camera poses to render full-body avatars from different views as well as zoom-in images of various body parts.
Specifically, we randomly sample camera poses from a spherical coordinate system with radius 3.5, elevation range $[-10^\circ,45^\circ]$, and $y$-axis field of view range $[-26^\circ, 45^\circ]$ for full-body renderings.
To encourage detailed body parts generation, we manipulate cameras to render zoom-in images for the face, back head, arms, upper body, and lower body. During training, we evenly sample different body parts and the full body renderings.

\vspace{-2mm}
\paragraph{Training.}
For each prompt, we optimize the avatar for 20000 iterations with the Adam optimizer. The learning rates for different learnable parameters discussed in Sec.~\ref{sec:optimization} of the main paper are listed in Table~\ref{tab::lr} below. We train the avatar in natural pose $\theta_{N}$ for 3000 iterations before introducing random pose $\theta_{A}$ sampled from the CMU motion capture database\footnote{\url{htp://mocap.cs.cmu.edu/}} using the SMPL-X parameters from AMASS~\cite{amass}. Starting from the 5000th iteration, we manipulate cameras to render zoom-in images for specific body parts (\eg, face, hands, upper body, \etc) to facilitate learning intricate detail in these parts.
The total training takes approximately 3 hours for each avatar on an NVIDIA RTX 3090Ti.

\begin{table}[!ht]
\centering
\footnotesize
\resizebox{\linewidth}{!}{ 
\begin{tabular}{cc}
\hline
Parameter & Learning rate \\
\hline
Gaussian local positions $\{p_k^i\}$ & 0.00016\\
Gaussian attribute field $\mathcal{H}_\phi$ & 0.001\\
SDF $\mathcal{S}_\psi$ & 0.0001\\
opacity kernel parameters $\{\gamma, \lambda\}$ & 0.001\\
primitive motion corrective networks $\delta P_\omega, \delta R_\omega, \delta S_\omega$ & 0.0001\\
the SMPL-X shape parameters $\beta$ & 0.0003\\
\hline
\end{tabular}
}
\vspace{-2mm}
\caption{\textbf{Learning rates for different parameters.}}
\label{tab::lr}
\vspace{-5mm}
\end{table}

\begin{figure*}[!ht]
    \centering
    \includegraphics[width=0.99\textwidth]{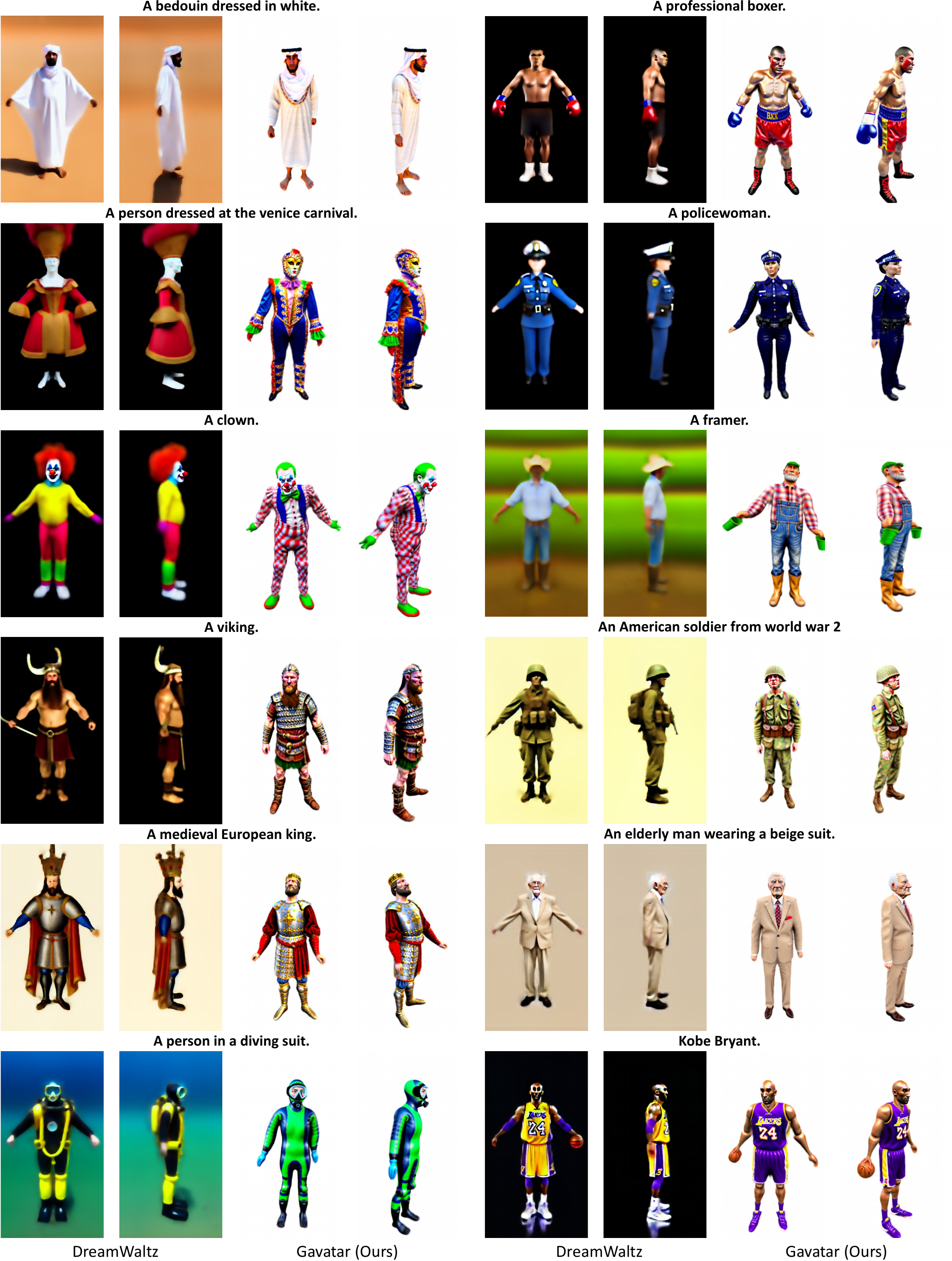}
    \vspace{-3mm}
    \caption{More comparisons with DreamWaltz~\cite{huang2023dreamwaltz}.}
    \label{fig::dreamwaltz}
    \vspace{-3mm}
\end{figure*}

\begin{figure*}[!ht]
    \centering
    \includegraphics[width=0.99\textwidth]{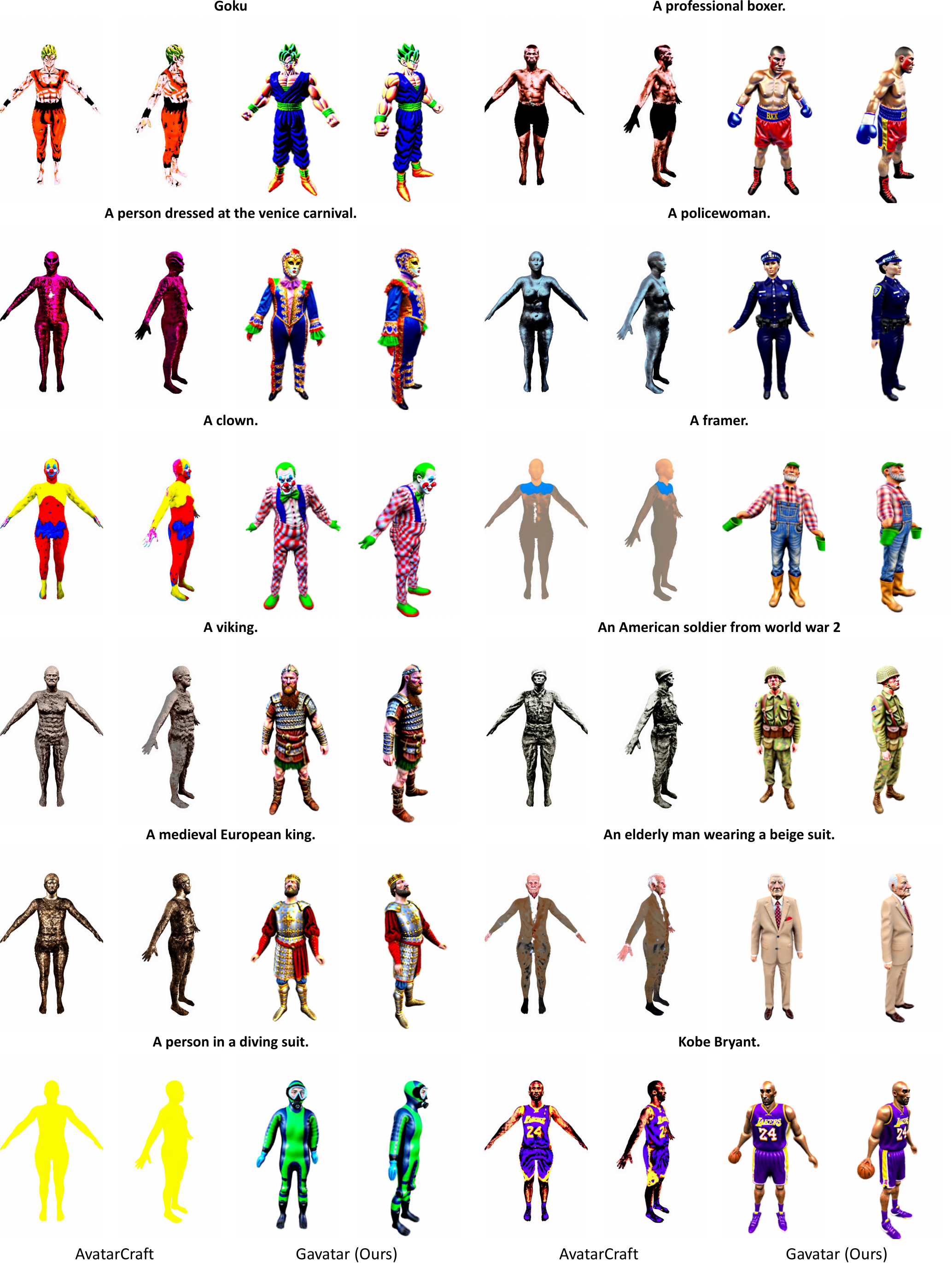}
    \vspace{-3mm}
    \caption{More comparisons with AvatarCraft~\cite{jiang2023avatarcraft}.}
    \label{fig::avatarcraft}
    \vspace{-3mm}
\end{figure*}

\begin{figure*}[!ht]
    \centering
    \includegraphics[width=0.99\textwidth]{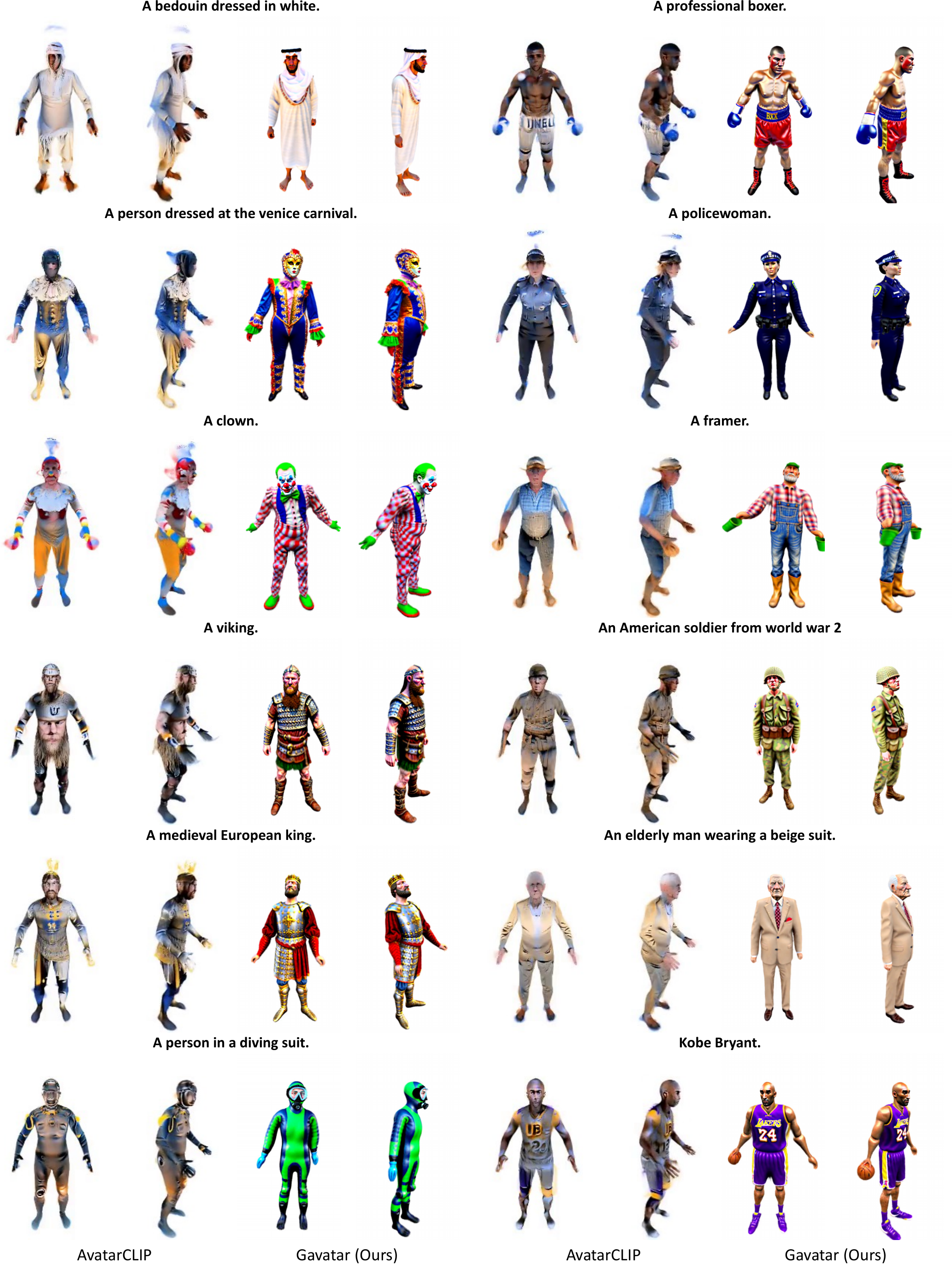}
    \vspace{-3mm}
    \caption{More comparisons with AvatarCLIP~\cite{hong2022avatarclip}.}
    \label{fig::avatarclip}
    \vspace{-3mm}
\end{figure*}

\begin{figure*}[!ht]
    \centering
    \includegraphics[width=0.99\textwidth]{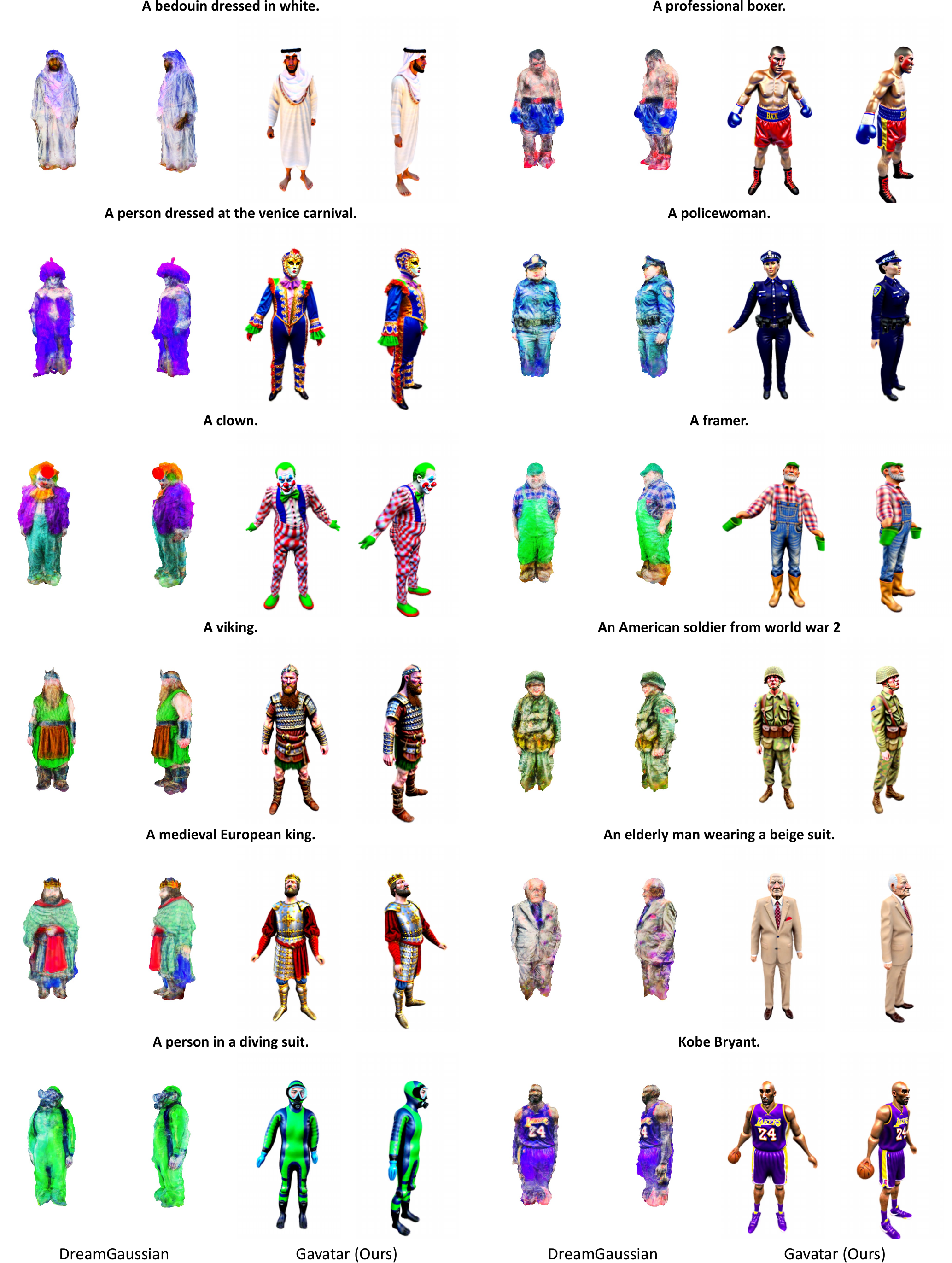}
    \vspace{-3mm}
    \caption{More comparisons with DreamGaussian~\cite{tang2023dreamgaussian}.}
    \label{fig::dreamgaussian}
    \vspace{-3mm}
\end{figure*}

\begin{figure*}[!ht]
    \centering
    \includegraphics[width=0.99\textwidth]{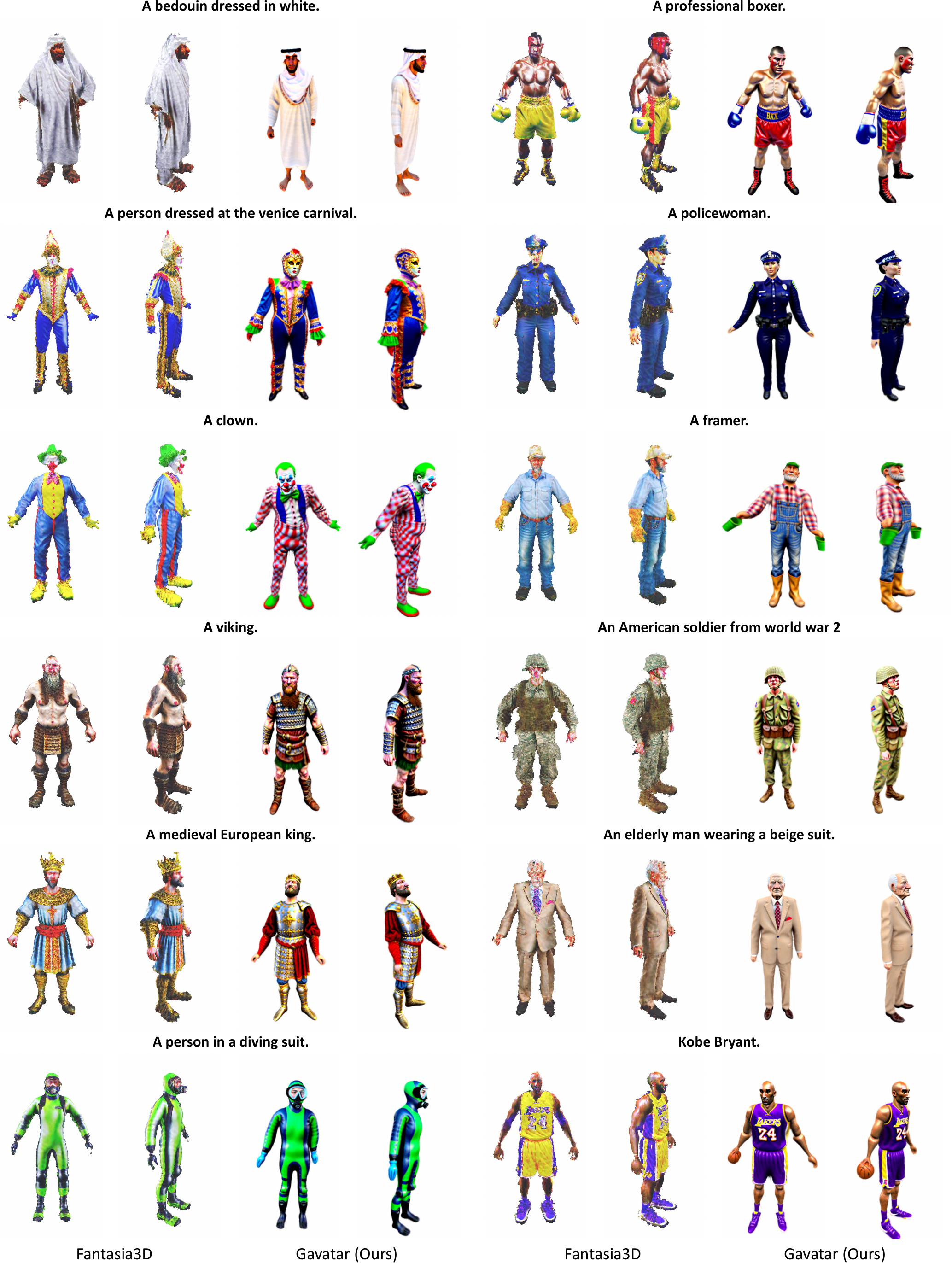}
    \vspace{-3mm}
    \caption{More comparisons with Fantastia3D~\cite{chen2023fantasia3d}.}
    \label{fig::fantastia}
    \vspace{-3mm}
\end{figure*}

\begin{figure*}[!ht]
    \centering
    \includegraphics[width=0.99\textwidth]{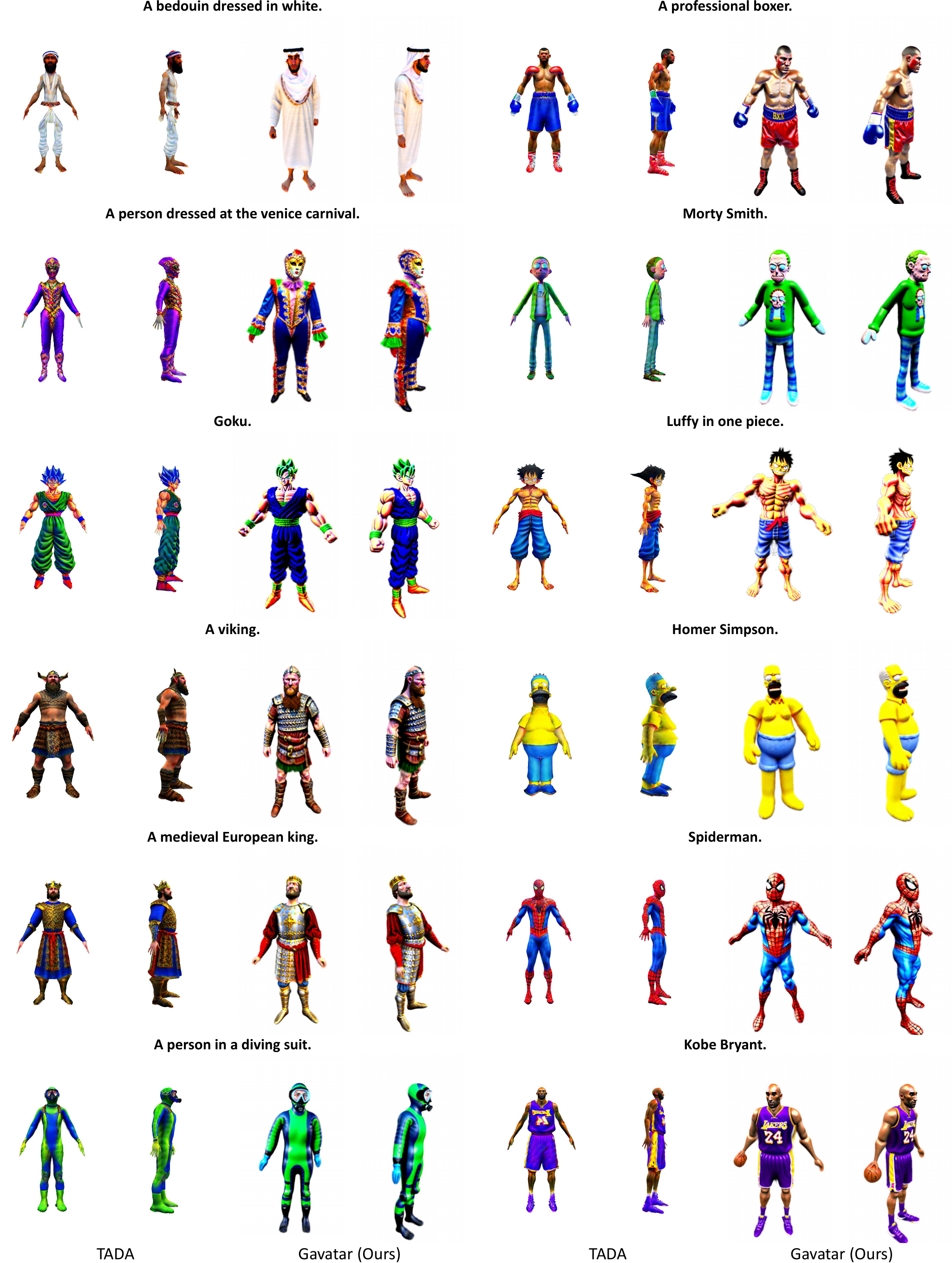}
    \vspace{-3mm}
    \caption{More comparisons with TADA~\cite{liao2024tada}.}
    \label{fig::tada}
    \vspace{-3mm}
\end{figure*}

\begin{figure*}[!ht]
    \centering
    \includegraphics[width=0.99\textwidth]{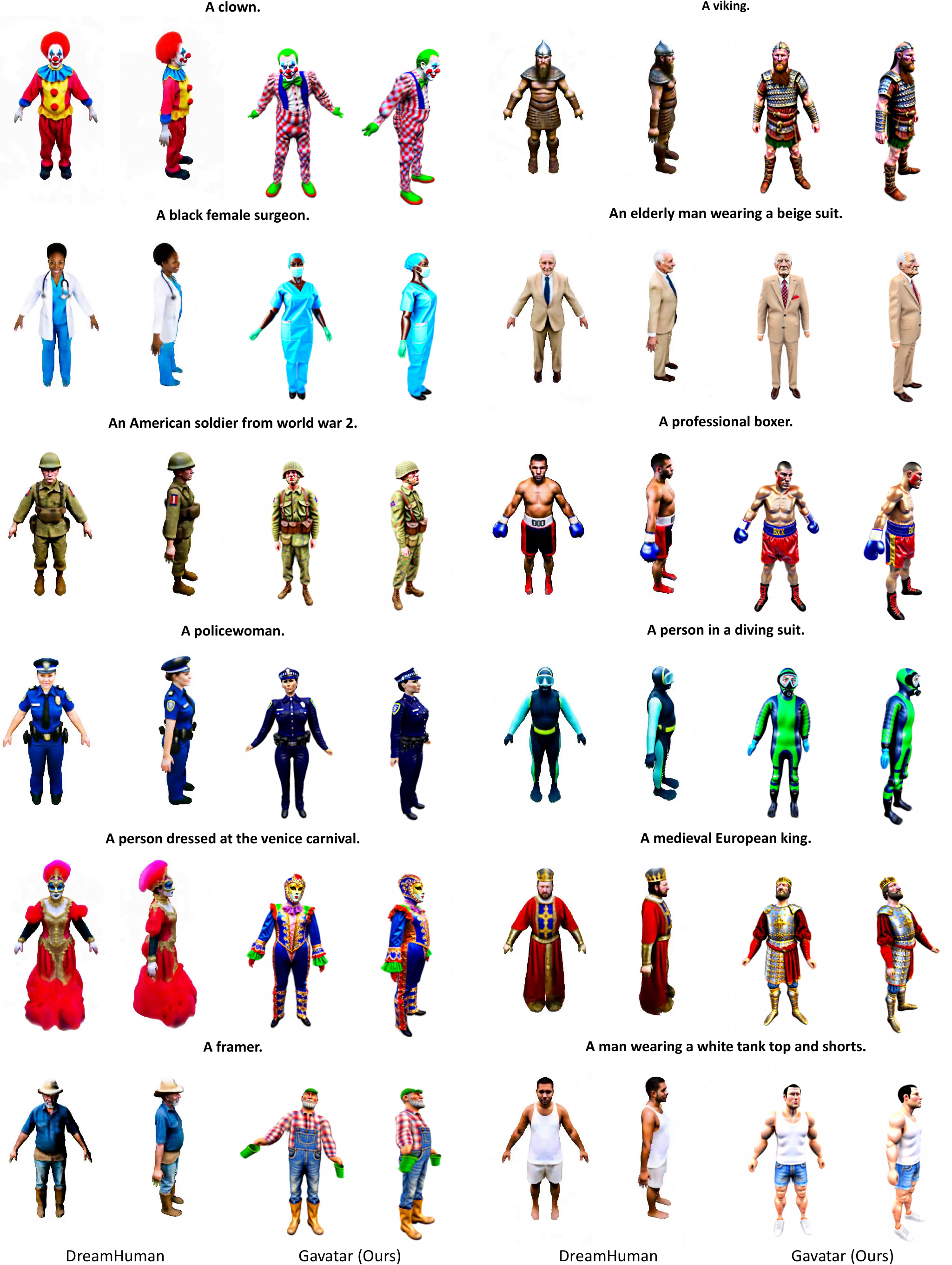}
    \vspace{-3mm}
    \caption{More comparisons with DreamHuman~\cite{kolotouros2023dreamhuman}.}
    \label{fig::dreamhuman}
    \vspace{-3mm}
\end{figure*}

\begin{figure*}[!ht]
    \centering
    \includegraphics[width=0.99\textwidth]{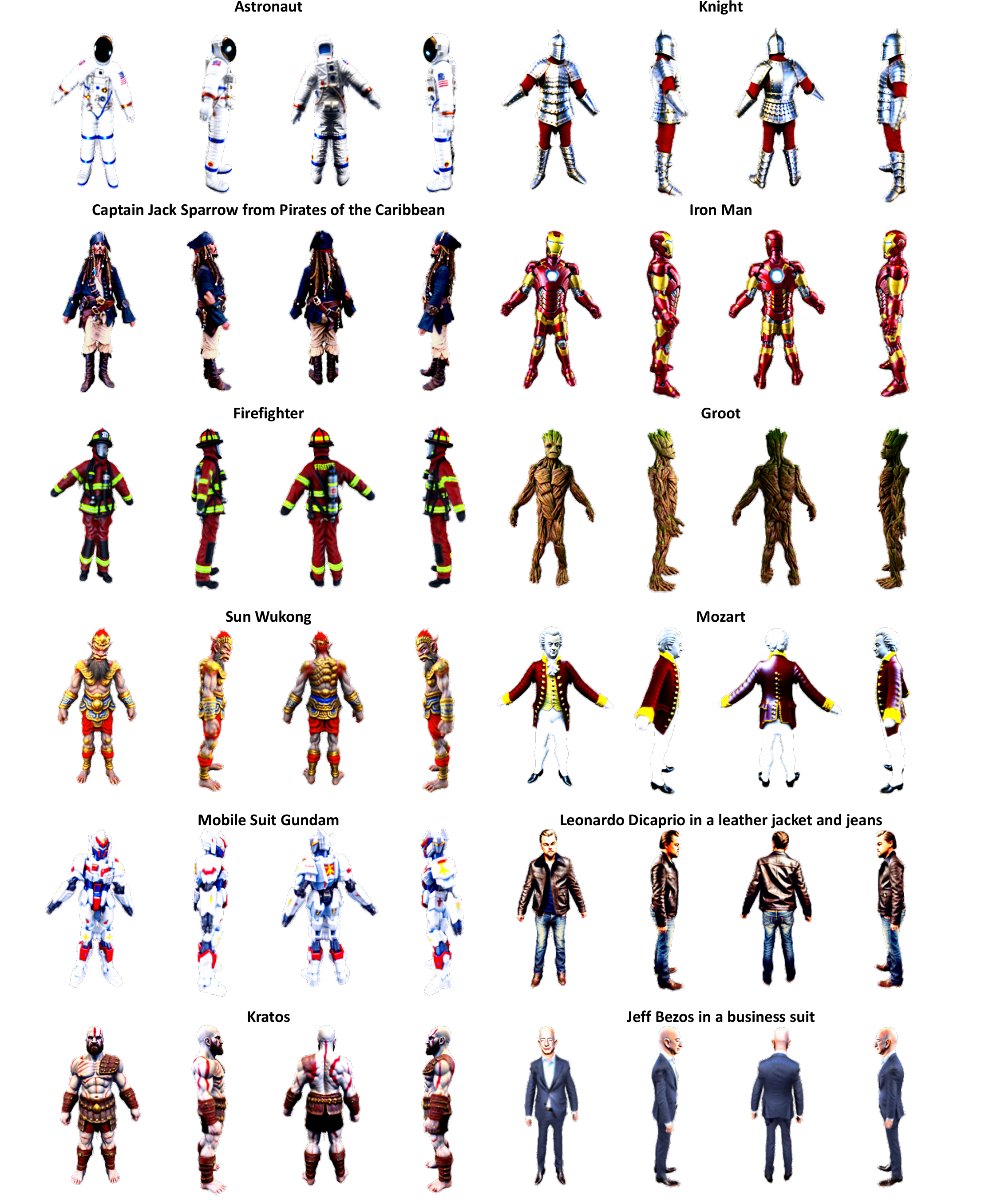}
    \vspace{-3mm}
    \caption{More results by GAvatar.}
    \label{fig::more_result1}
    \vspace{-3mm}
\end{figure*}

\begin{figure*}[!ht]
    \centering
    \includegraphics[width=0.99\textwidth]{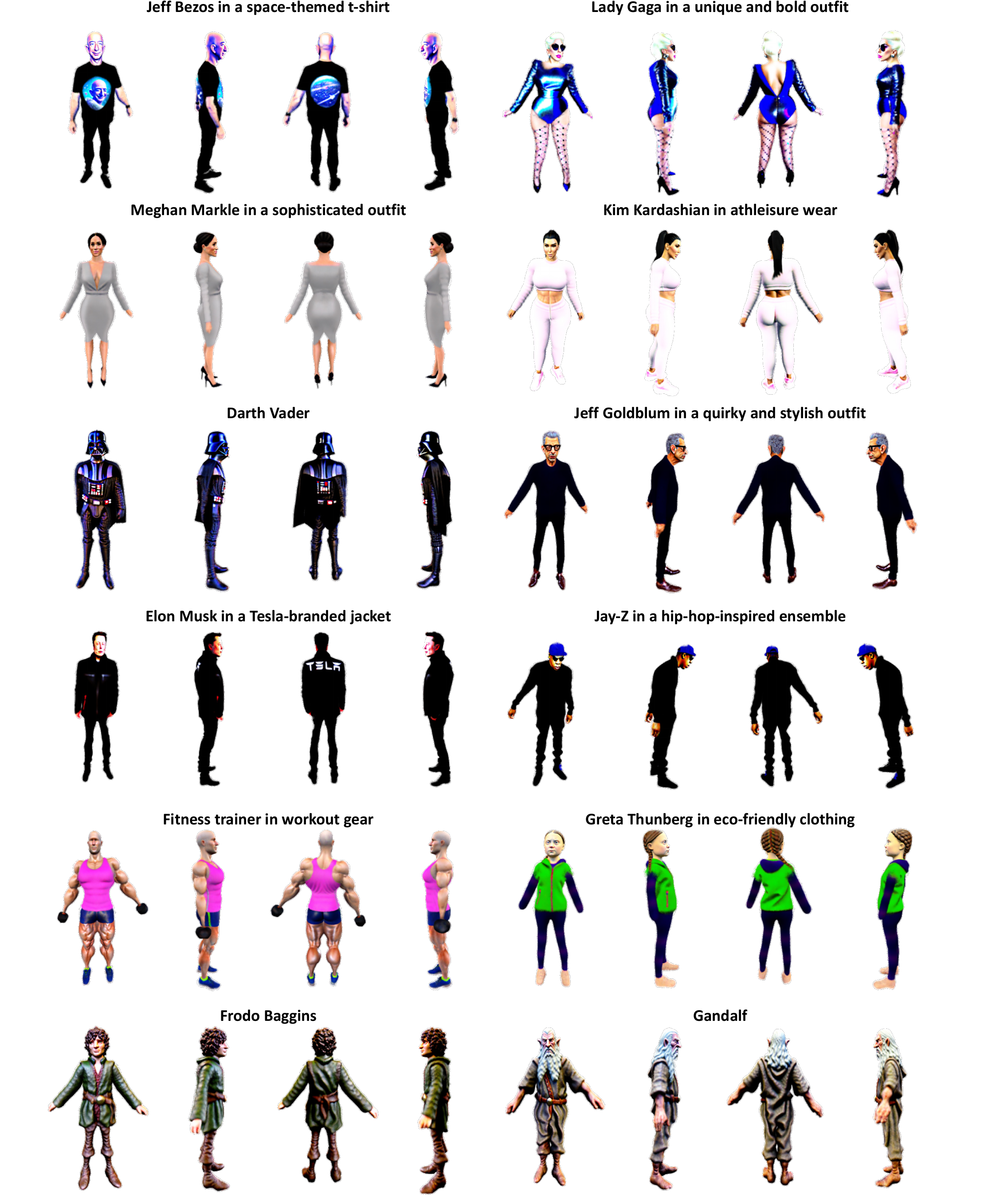}
    \vspace{-3mm}
    \caption{More results by GAvatar.}
    \label{fig::more_result2}
    \vspace{-3mm}
\end{figure*}

\vspace{-2mm}
\paragraph{Network architecture.} For the implicit Gaussian attribute field discussed in Sec.~\ref{sec:representation} in the main paper, we adopt a hash-encoded feature grid with $8$ levels, where the base resolution is $16\times 16\times 16$. The feature grid is followed by three MLP layers that output a $55$-dim vector including the scaling, rotation, and spherical harmonics features of the 3D Gaussian.
For the SDF discussed in Sec.~\ref{sec:sdf} in the main paper, we utilize a similar design as the Gaussian attribute field. Specifically, we use another hash-encoded feature grid with $16$ levels and a base resolution of $16\times 16\times 16$. The feature grid is followed by three MLP layers that output the SDF value of the 3D Gaussian, which is then converted to its opacity value using the opacity kernel $\mathcal{\kappa}$.
During training, we initialize each primitive with 64 Gaussians lying on a $4\times 4\times 4$ grid within the primitive and use the densification process (see Sec.~\ref{sec:optimization} in the main paper) to adaptively change the total Gaussian number as discussed in~\cite{kerbl3Dgaussians}. We also pretrain the Gaussian implicit fields to have an initial scale of 4mm in the world coordinate system.

\section{Additional Baseline Comparison}
\label{sec::supp_baseline}
\vspace{-1mm}
We provide additional qualitative comparisons with DreamWaltz~\cite{huang2023dreamwaltz}, AvatarCraft~\cite{jiang2023avatarcraft}, AvatarCLIP~\cite{hong2022avatarclip}, DreamGaussian~\cite{tang2023dreamgaussian}, Fantasia3D~\cite{chen2023fantasia3d}, TADA~\cite{liao2024tada} and DreamHuman~\cite{kolotouros2023dreamhuman} in Fig.~\ref{fig::dreamwaltz}, \ref{fig::avatarcraft}, \ref{fig::avatarclip}, \ref{fig::dreamgaussian}, \ref{fig::fantastia}, \ref{fig::tada} and \ref{fig::dreamhuman}, respectively. We note that DreamWaltz, DreamGaussian, TADA and DreamHuman are all concurrent text-to-3D avatar works. To ensure the best performance of the baselines, we use publicly available code and default hyper-parameters for each baseline except for DreamHuman, whose code is not available yet. Thus, we compare with the avatars downloaded from the project website\footnote{\url{htps://dream-human.github.io/}}.
Overall, our method is not only more robust to various prompts, but also shows more intricate and realistic details compared to all the baseline methods.

\section{Additional Qualitative Results}
\label{sec::supp_qualitative}
\vspace{-1mm}
We showcase more characters generated by GAvatar in Fig.~\ref{fig::more_result1} and~\ref{fig::more_result2}, demonstrating the robustness and generalization of the proposed method.

\section{User Study Prompts}
\label{sec::supp_prompts}
\vspace{-1mm}
For fair comparisons, we use the following 24 prompts commonly used by various baselines in the user study.

\begin{small}
\begin{verbatim}
A professional boxer.
Morty Smith.
A person in a diving suit.
An American soldier from World War 2.
Goku.
Rick Sanchez.
A person dressed at the Venice carnival.
A medieval European king.
An elderly man wearing a beige suit.
Kobe Bryant.
A man wearing a white tank top and shorts.
A policewoman.
A black female surgeon.
A viking.
Oprah Winfrey.
A bedouin dressed in white.
A framer.
A clown.
Jane Goodall.
Homer Simpson.
Kristoff in Frozen.
Luffy in one piece.
Spiderman.
Jeff Bezos.
\end{verbatim}
\end{small}